\documentclass[conference]{IEEEtran}
\usepackage{cite,amsmath,amssymb,graphicx,booktabs,microtype,url}
\usepackage[hidelinks]{hyperref}
\usepackage[table]{xcolor}
\usepackage{tikz}
\usetikzlibrary{arrows.meta,positioning,calc,fit,backgrounds}

\begin{document}

\title{How Hackable Is Your Speech Quality Metric?\\A Corrected Protocol, a Benchmark, and What Patching Buys}

\author{\IEEEauthorblockN{Ali Alavi}
\IEEEauthorblockA{Dept.\ of Computer Science and Engineering\\
The Ohio State University\\
alavibajestan.1@osu.edu}
\and
\IEEEauthorblockN{Donald S. Williamson}
\IEEEauthorblockA{Dept.\ of Computer Science and Engineering\\
The Ohio State University\\
williamson.413@osu.edu}}
\maketitle

\begin{abstract}
Speech quality predictors are increasingly used as rewards, yet no agreed measure of their
hackability exists. The usual measurement has two flaws. First, the perturbation reaches the predictor through a
processing chain --- here a neural codec --- that shifts the score on its own, which scoring against the raw input
charges to the attack. Referencing the unperturbed round trip instead changes measured hackability by up
to a factor of four ($0.31$ to $0.08$ for one defence). Second, one trained attacker is a sample, not a
measurement: five attackers differing only in random seed reach success rates from $0.00$ to $0.38$ against one
fixed predictor, so a defence claim needs the worst case over several. Under this protocol, four published
predictors differ widely: NISQA is hacked on 90\% of utterances, SSL-MOS on 21\%, DNSMOS on 14\% and UTMOS on
6\%. We then audit a closed attack--detect--patch loop. It hardens the predictor only in its own attack
space, by less than the spread between attackers; a random-perturbation baseline matches it; and it costs up
to 0.30 system SRCC out of domain. Enhancers post-trained against patched predictors hack them far less (PESQ
$-0.03$ versus $-0.23$). Code, preregistration and run outputs are released.
\end{abstract}


\section{Introduction}
\label{sec:intro}

Automatic mean-opinion-score (MOS) predictors are both the standard evaluation instrument for speech
generation~\cite{asonitis2026limits,huang2026attacking} and, increasingly, the \emph{reward} that trains
it~\cite{kumar2025rlhf,wang2026flowsegrpo,berdoz2026gspo}. Optimising such a metric degrades the perceptual
quality it was built to measure: an enhancer tuned for PESQ reaches a state-of-the-art 3.82 PESQ yet scores % lint:ok (cited from prior work)
very poorly in a listening test~\cite{deoliveira2024pesqetarian}. The predictors can also be attacked directly. An
attack adds a small perturbation --- to the waveform, or to a neural codec's latent representation of it ---
and asks how far the predicted score moves; a white-box attack shapes it with the predictor's gradients, a
black-box attack sees only the returned score. White-box attacks on UTMOS start
from clean speech, degrade it severely, and still hold its predicted MOS at the clean
value~\cite{huang2026attacking}: a high predicted score is no evidence of high quality. What such an
attack does not do is push the score \emph{above} the score of the speech it started from --- and that is
exactly what a generator trained against the predictor is rewarded for, since every point it adds is reward it
collects. A practitioner choosing which predictor to reward with therefore has no evidence to go on: predictors
are picked for their published correlation with human ratings, for precedent, or for what a toolkit ships ---
none of which says how much score an optimiser can extract without improving the audio.

There is no protocol for measuring how hackable a predictor is, and no study that puts the predictors people
actually use in front of the \emph{same} attacker, at the \emph{same} perturbation strength, under the
\emph{same} rule for what counts as a hack. Published attacks differ in all three. Lin et
al.~\cite{lin2022robustness} drive DNSMOS to a fixed target score under a $-30$\,dB peak bound and treat any % lint:ok (cited from prior work)
change in the prediction as an attack; EventTrojan~\cite{ren2023eventtrojan} poisons $3\%$ of a predictor's
training data with an inserted $0.2$--$0.5$\,s dropout and counts a score landing within $0.5$ MOS of the
attacker's target; the UTMOS attacks~\cite{huang2026attacking} impose no norm bound at all and report a
PESQ--UTMOS trade-off over 30 utterances. Their numbers therefore do not share an % lint:ok (cited from prior work)
axis: a predictor can look robust because it is, or merely because it was attacked more gently.

The same failure sits in every pipeline that puts a learned quality model inside an optimisation loop. Speech
enhancement is now routinely post-trained against DNSMOS or NISQA
rewards~\cite{wang2026flowsegrpo,wang2025sqase}: any perturbation that raises the reward without improving the
speech is a shortcut, and RL is a search that finds shortcuts, so a deployed system can degrade on the very axis
it reports improving. Text-to-speech and voice conversion use the same predictors to filter training
corpora~\cite{he2024emilia,ma2024wenetspeech4tts}, pick checkpoints and
hyper-parameters~\cite{peirolilja2022monitoring,nakata2024utduss} and rank challenge
submissions~\cite{chang2024discrete}, so a hackable predictor quietly promotes the wrong system. In hearing aids these predictors stand in for
listeners who cannot easily A/B test their own device, so a metric an inaudible perturbation lifts by a full MOS
point gives no assurance that a fitting which scores better sounds better. In each case the practitioner's only
lever is which predictor to trust, and that choice is currently made blind.

This paper supplies the missing protocol and the missing comparison. A hack is a score that rises
\emph{because of the perturbation}, so the rule's reference is the processing chain's own output, not the raw
input, which folds in how the predictor scores the codec itself ($-0.10$ to $+0.14$ MOS across predictors, up to
$+0.95$ across defences). And because an attacker is itself trained, one attacker is one sample: five attackers
differing only in random seed span success rates from $0.00$ to $0.38$ against one fixed predictor. Fixing both,
we benchmark four widely used predictors at one perturbation size and find a spread of more than an order of
magnitude, the most hackable --- NISQA, commonly used as an RL reward --- hacked on $90\%$ of utterances.

We then apply the protocol to the defence the diagnosis suggests: a closed attack--detect--patch loop, the
speech counterpart of Adv-RM~\cite{bukharin2025advrm}, whose modality gap we cannot cross to run it as a
baseline. Correctly measured, it hardens only the space it trains on, by less than the spread between attackers,
and a baseline with no RL and no detector matches it.

\noindent\textbf{Contributions.} (i) A measurement protocol --- reference the hack rule to the processing
chain's output, report the worst case over $k$ trained attackers --- with evidence that both corrections change
conclusions (Sec.~\ref{sec:results}). (ii) The first benchmark of common predictors at one perturbation size and
one detector (Table~\ref{tab:bench}, Fig.~\ref{fig:results}a). (iii) An audit of closed-loop patching: narrow
hardening, negative transfer, a trivial baseline that matches it, and the out-of-domain cost
(Table~\ref{tab:e1}). (iv) Downstream evidence on reward hacking of a small enhancement model
(Table~\ref{tab:e3}). All data, models and analysis scripts are released.

\section{Related work}
\label{sec:related}
\noindent\textbf{Quality metrics can be gamed.} Beyond the PESQetarian~\cite{deoliveira2024pesqetarian}, % lint:ok (cited from prior work)
FlowSE-GRPO observes implausible DNSMOS gains under single-metric
optimisation~\cite{wang2026flowsegrpo}, and composite rewards are the usual
remedy~\cite{berdoz2026gspo,wang2025sqase}. Asonitis et al.~\cite{asonitis2026limits} diagnose hacking --- the
optimised score rises as an independent panel falls --- but leave the predictor untouched.

\noindent\textbf{Attacking quality predictors.} None of the attacks of Sec.~\ref{sec:intro} proposes a
defence. Lin et al.~\cite{lin2022robustness} adversarially train DNSMOS once against waveform perturbations,
without a hack detector or a downstream generator; RRPO~\cite{wang2026rrpo} hardens a TTS reward model in one
pass, as does gradient-norm regularisation for image quality models~\cite{liu2024iqadefense}.

\noindent\textbf{Closing the loop.} Adv-RM~\cite{bukharin2025advrm} applies the same recipe to LLM reward models. Related ideas alternate policy and reward-model
updates~\cite{cheng2024apo,hong2025cooper,wolf2025iterated}, ensemble reward
models~\cite{coste2024ensembles,eisenstein2024helping}, harden agent benchmarks~\cite{zhong2026hackerfixer} or
post-train adversarially for music~\cite{wu2026gapt}.

\begin{figure}[t]
\centering
\resizebox{\columnwidth}{!}{%
\begin{tikzpicture}[
  font=\scriptsize, >={Stealth[length=1.6mm]},
  box/.style={draw, rounded corners=1pt, align=center, inner sep=2pt, minimum height=4.8mm},
  sig/.style={box, fill=blue!6}, mod/.style={box, fill=orange!12},
  rule/.style={box, fill=green!8, align=left},
  lbl/.style={font=\tiny, inner sep=1.2pt}]

% ---- attack chain: gaps sized to the edge labels they carry ----
\node[sig] (x) {clean\\speech $x$};
\node[mod, right=6mm of x] (enc) {codec\\encoder};
\node[mod, right=7mm of enc] (pol) {policy $\pi_\theta$\\(GRPO)};
\node[mod, right=15mm of pol] (dec) {codec\\decoder};
\node[sig, right=6mm of dec] (y) {attacked\\output $y$};

\draw[->] (x) -- (enc);
\draw[->] (enc) -- node[lbl, above] {$z$} (pol);
\draw[->] (pol) -- node[lbl, above] {$z\!+\!\delta$}
                   node[lbl, below, align=center] {$\|\delta\|_{\mathrm{rms}}$\\$=\epsilon$} (dec);
\draw[->] (dec) -- (y);

% ---- zero-perturbation bypass: the corrected reference, routed above the row ----
\draw[->, dashed] (enc.north) -- ++(0,5mm)
  -- node[lbl, above] {$\delta\!=\!0$: round trip $\hat{x}$} ($(dec.north)+(0,5mm)$) -- (dec.north);

% ---- scoring ----
\node[mod, below=7mm of pol] (R) {predictor $R_{k-1}$};
\draw[->] (y) |- (R);
\draw[->] (R) -- node[lbl, right=0.4mm] {reward $R(y)$} (pol);

% ---- detector ----
\node[rule, below=5mm of R, xshift=14mm] (det)
  {\textbf{hack?} \; $R(y)-R(x) > \tau$\\ \textbf{and} one of ESTOI, SI-SDR, WER,\\ SpeechBERTScore worse than $\hat{x}$};
\draw[->] (R.south) -- (det.north -| R.south);

\node[sig, left=9mm of det] (pair) {pair\\$\hat{x} \succ y$};
\draw[->] (det.west) -- node[lbl, above] {if flagged} (pair.east);

\node[mod, below=4mm of pair] (patch) {patch $R_k$:\ MSE on BVCC\\$+$ ranking loss over replay buffer};
\draw[->] (pair) -- (patch);
\draw[->] (patch.west) -- ++(-4mm,0) |- node[lbl, pos=0.78, above] {round $k\!+\!1$} (R.west);
\end{tikzpicture}}
\caption{The attack--detect--patch loop. A GRPO policy perturbs codec latents at a fixed size; an output is
flagged only if its score rises above that of the zero-perturbation round trip $\hat{x}$ (dashed) \emph{and} an
independent panel degrades relative to $\hat{x}$. Flagged pairs patch the predictor; the next round retrains the
attacker.}
\label{fig:method}
\end{figure}
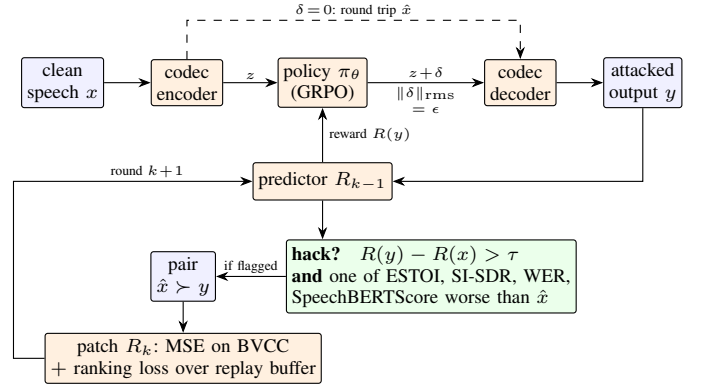

\section{Method}
\label{sec:method}
Let $R$ be a mean-opinion-score (MOS) predictor mapping a waveform to a scalar. The loop (Fig.~\ref{fig:method}) has three parts.

\noindent\textbf{Attacker.}
Speech $x$ is encoded to a continuous latent $z=\mathrm{Enc}(x)$ by a frozen neural codec~\cite{defossez2023encodec}. A policy
$\pi_\theta$ proposes an action $a\sim\mathcal{N}(\mu_\theta(z),\sigma^2)$ in per-channel standardised latent
units; the perturbation $\delta=\mathrm{expand}(a)$ is rescaled to a fixed root-mean-square size $\epsilon$,
and the attacked signal is $y=\mathrm{Dec}(z+\delta\odot s_z)$ with $s_z$ the per-channel latent standard
deviation. The policy is trained with GRPO~\cite{shao2024grpo}: for each utterance we draw $G$ actions, score them with the frozen
target predictor, and use group-normalised advantages $A_{ij}=(r_{ij}-\bar r_i)/(\mathrm{std}_j r_i+\eta)$ with
$r=R(y)$, optimised with a clipped surrogate. The action is time-invariant (one $C$-dimensional offset per
utterance), which keeps the policy-gradient variance tractable: a frame-rate action space did not learn at any
size we tried. The attacker needs only forward scores, so any predictor can be a target.

\noindent\textbf{Label-free hack detection.}
An attacked output $y$ is flagged as a \emph{hack} when both rules hold:
(i) \emph{ceiling rule} --- the predictor scores the attacked signal more than $\tau$ MOS above the codec round
trip $\hat{x}=\mathrm{Dec}(z)$, i.e. $R(y)-R(\hat{x})>\tau$; and
(ii) \emph{panel rule} --- at least one independent measure is worse for $y$ than for $\hat{x}$ by more than a
per-measure margin: ESTOI, SI-SDR, SpeechBERTScore~\cite{saeki2024speechbertscore} (all against $x$),
or the word error rate of Whisper transcripts of $y$ against those of $x$.
Referencing both rules to $\hat{x}$ rather than to the raw input $x$ is what isolates the effect of the
perturbation from the codec's own round-trip loss; the loop as run used $x$ in the ceiling rule, and every
number we report is recomputed against $\hat{x}$ (Sec.~\ref{sec:results}). No human labels are used. The fraction of utterances an attacker gets flagged this
way is its \emph{attack success rate} (ASR), our measure of hackability.

\noindent\textbf{Patch.}
Every flagged hack contributes a preference pair $\hat{x}\succ y$, and the predictor is fine-tuned on its
original regression loss plus a margin ranking loss $\max(0, m-(R(\hat{x})-R(y)))$ over a replay buffer holding
every previous round's hacks. Using $\hat{x}$ rather than $x$ as the preferred side prevents a ``codec
artefact'' shortcut; an exploratory ablation ($n\!=\!1$) confirms it (in-loop ASR $0.010$ by round~3 against
$0.172$ with $x$). The three steps repeat for $K$ rounds, giving $R_1,\dots,R_K$.

\noindent\textbf{Multi-space variant.}
The same loop can draw its pairs from several attack spaces at once. In addition to the codec latent space above
we use (i) a second codec (DAC~\cite{kumar2023dac}, quantised latent) and (ii) white-box waveform PGD that
maximises $R$ under $\|\delta\|_\infty\le 0.002$. All flagged hacks share one buffer.

\section{Experimental setup}
\label{sec:setup}
\noindent\textbf{Predictor and data.} $R_0$ is SSL-MOS~\cite{cooper2022generalization}: a wav2vec\,2.0 base % lint:ok
encoder with mean pooling and a linear head, trained with MSE on the BVCC~\cite{cooper2021bvcc} training % lint:ok
split (4\,974 utterances) % lint:ok (dataset size)
for 8\,000 steps (AdamW, $2\!\times\!10^{-5}$ encoder / $10^{-3}$ head, batch 8, 6\,s crops), selecting the best % lint:ok
validation-MSE checkpoint. Human agreement is evaluated on 13 test sets: BVCC test, the VoiceMOS'22
out-of-domain set~\cite{huang2022voicemos,wu2019blizzard}, three NISQA sets~\cite{mittag2021nisqa}, % lint:ok
SOMOS~\cite{maniati2022somos}, TMHINT-QI~\cite{chen2022inqss}, four VoiceMOS'23
tracks~\cite{cooper2023voicemos} and the URGENT'24/'25 quality-assessment % lint:ok (dataset name)
sets~\cite{zhang2024urgent,saijo2025urgent}.

\noindent\textbf{Attacker and detector.} The attack space is the continuous EnCodec-24\,kHz latent % lint:ok
(128 channels at 75\,Hz); the policy is a four-layer 1-D convolutional network (256 hidden units) with a % lint:ok
per-channel learned $\sigma$, trained for 2\,000 GRPO steps ($G=8$, batch 8, 3\,s crops, lr $10^{-3}$, % lint:ok
clip 0.2, two inner epochs). $\epsilon=0.5$ was selected on validation alone (Sec.~\ref{sec:results}). The detector uses $\tau=0.3$ MOS with margins of 0.02 ESTOI, 1\,dB SI-SDR, 0.02

\noindent\textbf{Patching and baselines.} Each round fine-tunes for 1\,500 steps with equal MSE and ranking % lint:ok
batches (margin $m=0.5$, weight 1) and $K=3$ rounds. Three defences share the 4\,500-step budget and loss, % lint:ok
differing in where the rejected side comes from: Gaussian noise pairs (SNR 5--30\,dB), waveform-PGD % lint:ok
pairs, and random latent perturbations at the same $\epsilon$ (no RL, no detector). PGD uses
$\|\delta\|_\infty\!\le\!0.002$ with 50 steps at evaluation but 10 steps inside the multi-space loop and 5 in the % lint:ok
PGD-pair baseline, so those defences meet a weaker attacker than they are tested against; training is
step-matched but not pair-matched (replay buffer 5.8k to 8.9k pairs against 19.9k random pairs, % lint:ok
favouring the baseline).

\noindent\textbf{Downstream (E3).} A 2.6\,M-parameter STFT mask enhancer is pre-trained on VoiceBank-DEMAND and % lint:ok
post-trained with GRPO for 2\,500 steps, the policy adding per-frame offsets on 32 ERB bands with no KL % lint:ok
penalty, so only the reward model limits hacking. Evaluation uses the 824-utterance test set. % lint:ok

\noindent\textbf{Protocol.} Seeds, primary metrics and decision rules for E1--E3 were preregistered before any
test-label contact, with one amendment (attack action space and the $\epsilon$ rule) made on validation evidence
before the main runs; the multi-space variant and the ablations are exploratory, and the corrected ceiling
reference is a post-hoc re-analysis of stored outputs. Results are mean\,$\pm$\,sd over $n$ seeds.

\begin{figure*}[t]
\centering
\includegraphics[width=\textwidth]{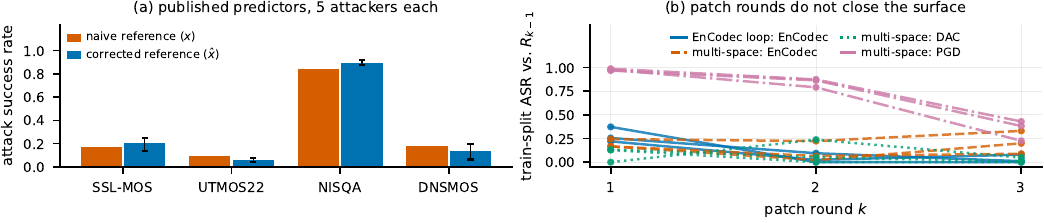}
\caption{(a) Benchmark of four common predictors at one perturbation size under the naive ($x$) and corrected
($\hat{x}$) ceiling references; error bars span the five attackers, and the reference moves DNSMOS and NISQA in
\emph{opposite} directions. (b) In-loop attack success rate (BVCC train, corrected reference) of the round-$k$
attacker against the predictor patched in round $k\!-\!1$, one line per seed: the single-space loop converges in
its own space; under multi-space training DAC and PGD fall while the EnCodec attacker stays flat ($+0.01$).}
\label{fig:results}
\end{figure*}

\section{Results}
\label{sec:results}

\begin{table}[t]
\centering
\caption{\textbf{Hackability of common predictors} at one size ($\epsilon\!=\!0.5$, EnCodec latents), five
independently trained attackers each, 500 BVCC-test utterances; ``worst'' is the largest of the five.}
\label{tab:bench}
\footnotesize
\setlength{\tabcolsep}{3pt}
% generated by analysis/paper_assets.py
\begin{tabular}{l c c c c}
\toprule
Predictor & offset & mean ASR & worst ASR & mean gain \\
\midrule
SSL-MOS (BVCC) & -0.10 & 0.205 & 0.252 & +0.191 \\
UTMOS22 & -0.07 & 0.057 & 0.076 & +0.144 \\
NISQA & +0.14 & 0.896 & 0.916 & +0.954 \\
DNSMOS OVRL & +0.02 & 0.136 & 0.200 & +0.124 \\
\bottomrule
\end{tabular}

\end{table}

% \begin{figure*}[t]
% \centering
% \includegraphics[width=\textwidth]{figs/results.pdf}
% \caption{(a) Benchmark of four predictors under the naive and corrected references; error bars span the five
% attackers. (b) In-loop attack success rate (BVCC train, corrected reference) of the round-$k$ attacker against
% the predictor patched in round $k\!-\!1$, one line per seed.}
% \label{fig:results}
% \end{figure*}

\noindent\textbf{The reference of the hack rule changes the answer.} A predictor that scores codec-processed
audio differently from raw audio shifts every raw-referenced measurement with it, and that offset is neither
negligible nor constant: $-0.10$ (SSL-MOS), $-0.07$ (UTMOS), $+0.02$ (DNSMOS) and $+0.14$
(NISQA) across predictors (Table~\ref{tab:bench}), and up to $+0.95$ across the defences of
Table~\ref{tab:e1}. The consequences are concrete: under the naive reference DNSMOS looks \emph{more}
hackable than it is ($0.183$ vs $0.136$), NISQA \emph{less} ($0.839$ vs $0.896$), and the multi-space defence
below flips from apparently catastrophic ($0.306$) to among the most robust ($0.082$).

\noindent\textbf{One attacker is not a measurement.} Retraining the attacker with different seeds against one
fixed predictor gives ASR from $0.000$ to $0.261$ on $R_0$ and $0.000$ to $0.382$ on $R_3$ (sd $0.130$ and
$0.180$) --- as large as every defence effect we measure. Between \emph{predictors}, by contrast, the spread is
small relative to the differences (NISQA $0.876$--$0.916$ against UTMOS $0.042$--$0.076$), so the benchmark
ranking is solid where defence comparisons are not. We therefore report the worst of five attackers.

\noindent\textbf{Common predictors differ by more than an order of magnitude (Table~\ref{tab:bench}).} At
one perturbation size, NISQA is hacked on $89.6\%$ of utterances for a mean gain of $+0.954$ MOS, SSL-MOS on
$20.5\%$, DNSMOS OVRL on $13.6\%$ and UTMOS on $5.7\%$; worst-case rates are $0.916$, $0.252$, $0.200$ and
$0.076$. As NISQA and DNSMOS are routinely used as RL rewards, the choice of reward is a
first-order robustness decision, and the ranking is not predictable from in-domain accuracy.

\begin{table*}[t]
\centering
\caption{\textbf{Auditing the loop.} Corrected-reference ASR per attack space (fresh attacker, BVCC test) and
agreement with human MOS (BVCC system SRCC; out-of-domain mean over 12 sets), mean$\pm$sd over 3 seeds.}
\label{tab:e1}
\scriptsize
\setlength{\tabcolsep}{3.5pt}
% generated by analysis/paper_assets.py
\begin{tabular}{l c c c c c c}
\toprule
& offset & \multicolumn{3}{c}{ASR by attack space} & \multicolumn{2}{c}{human agreement} \\
\cmidrule(lr){3-5}\cmidrule(lr){6-7}
Predictor & $R(\hat{x})\!-\!R(x)$ & EnCodec & DAC & PGD & BVCC & OOD \\
\midrule
$R_0$ (no defense) & -0.09$\pm$0.02 & 0.226$\pm$0.205 & 0.065$\pm$0.113 & 0.999$\pm$0.002 & 0.922$\pm$0.004 & 0.731$\pm$0.006 \\
$R_1$ & -0.06$\pm$0.03 & 0.050$\pm$0.048 & 0.161$\pm$0.033 & 1.000$\pm$0.001 & 0.911$\pm$0.009 & 0.690$\pm$0.014 \\
$R_2$ & 0.00$\pm$0.07 & 0.017$\pm$0.006 & 0.141$\pm$0.025 & 1.000$\pm$0.000 & 0.918$\pm$0.003 & 0.682$\pm$0.008 \\
$R_3$ (EnCodec loop) & -0.27$\pm$0.25 & 0.137$\pm$0.179 & 0.127$\pm$0.031 & 1.000$\pm$0.000 & 0.920$\pm$0.001 & 0.637$\pm$0.046 \\
$R_3^{\mathrm{multi}}$ (multi-space) & 0.95$\pm$0.16 & 0.082$\pm$0.107 & 0.016$\pm$0.026 & 0.841$\pm$0.089 & 0.914$\pm$0.003 & 0.552$\pm$0.015 \\
noise pairs & -0.09$\pm$0.01 & 0.270$\pm$0.107 & 0.230$\pm$0.028 & 1.000$\pm$0.000 & 0.917$\pm$0.004 & 0.722$\pm$0.013 \\
PGD pairs & -0.08$\pm$0.02 & 0.223$\pm$0.057 & 0.225$\pm$0.044 & 0.969$\pm$0.025 & 0.909$\pm$0.001 & 0.635$\pm$0.009 \\
random-latent pairs & -0.03$\pm$0.03 & 0.019$\pm$0.016 & 0.200$\pm$0.053 & 1.000$\pm$0.000 & 0.922$\pm$0.004 & 0.696$\pm$0.010 \\
\bottomrule
\end{tabular}

\end{table*}

\noindent\textbf{Patching hardens narrowly, and a trivial baseline matches it (Table~\ref{tab:e1}).} Three
rounds against the EnCodec attacker cut ASR from $0.226\!\pm\!0.205$ ($R_0$) to $0.050$ ($R_1$) and $0.017$
($R_2$), and the replayed round-1 attacker never succeeds. The effect is not monotone --- $K\!=\!3$ was
preregistered rather than selected, and $R_2$ dominates $R_3$ on both axes --- and the $R_0\!\to\!R_3$ difference
does not hold its sign across seeds ($-0.073$, $-0.039$, $+0.379$), so our preregistered H1, which passes under
the naive reference, fails under the corrected one. Pairs from \emph{random} latent perturbations at the same
size, with no RL and no detector, give the lowest in-space ASR of any defence ($0.019\!\pm\!0.016$). Since the
patch margin ($m\!=\!0.5$) exceeds the detection threshold ($\tau\!=\!0.3$) we swept $\tau\in[0.1,1.0]$: the
ordering is unchanged, though $R_3$ alone keeps a tail of large hacks (ASR $0.045$ at $\tau\!=\!1.0$;
90th-percentile score gap $0.651$ vs $0.359$ for $R_0$).

\begin{table}[!tb]
\centering
\caption{\textbf{Downstream GRPO enhancement.} Change against the supervised pre-trained enhancer on
VoiceBank-DEMAND test, per reward predictor. Mean$\pm$sd over 3 seeds ($R_2$: one seed).}
\label{tab:e3}
\scriptsize
\setlength{\tabcolsep}{3pt}
% generated by analysis/paper_assets.py
\begin{tabular}{l c c c}
\toprule
Measure & $R_0$ & $R_3$ & random-latent pairs \\
\midrule
$\Delta$ own reward & 0.263$\pm$0.010 & 0.269$\pm$0.028 & 0.334$\pm$0.051 \\
$\Delta$PESQ & -0.226$\pm$0.170 & -0.034$\pm$0.126 & 0.019$\pm$0.050 \\
$\Delta$ESTOI & -0.057$\pm$0.010 & -0.038$\pm$0.013 & -0.043$\pm$0.006 \\
$\Delta$SI-SDR (dB) & -6.1$\pm$1.4 & -5.8$\pm$1.4 & -5.7$\pm$0.6 \\
$\Delta$DNSMOS & -0.075$\pm$0.031 & -0.076$\pm$0.056 & -0.075$\pm$0.043 \\
$\Delta$NISQA & 0.335$\pm$0.186 & 0.300$\pm$0.072 & 0.468$\pm$0.059 \\
$\Delta$UTMOS & 0.018$\pm$0.006 & 0.057$\pm$0.030 & 0.076$\pm$0.011 \\
\bottomrule
\end{tabular}

\end{table}

\noindent\textbf{No transfer, and one real trade.} Against an unseen DAC attacker the EnCodec-patched predictors
are \emph{worse} than undefended ($0.065$ to between $0.13$ and $0.16$), and white-box PGD succeeds on nearly
every utterance for every EnCodec-trained defence. Only multi-space training changes this: it resists DAC
($0.016$) and is the one predictor PGD does not fully break ($0.841\!\pm\!0.089$), paying with the $+0.95$ MOS
reconstruction offset and the largest loss of human agreement. In-loop, DAC and PGD fall round by round while the
EnCodec attacker stays flat (Fig.~\ref{fig:results}b).

\noindent\textbf{Fidelity: ranking holds in domain, calibration and out-of-domain agreement do not.} The patched
predictors keep their BVCC system ranking ($R_3$ $0.920$ vs $R_0$ $0.922$, meeting the preregistered $-0.02$
margin) but utterance MSE rises $0.234\!\pm\!0.011\to0.272\!\pm\!0.022$, a calibration loss that matters for a
reward model. Out of domain the mean system SRCC falls from $0.731$ to $0.690$, $0.682$ and $0.637$ over rounds
one to three and to $0.552$ for $R_3^{\mathrm{multi}}$, worst on noisy speech (TMHINT-QI, where the $R_0$ and
$R_3$ bootstrap intervals do not overlap; TMHINT-QI and VoiceMOS'23 track~3 share a condition grid and are not
independent). A predictor taught that perturbed audio is bad seems to generalise that to genuinely noisy audio.

\noindent\textbf{Downstream, patched predictors are safer rewards --- and so is the baseline
(Table~\ref{tab:e3}).} Against $R_0$ the enhancer shows the textbook hack: reward $+0.263$ while PESQ falls
$-0.226$, ESTOI $-0.057$ and SI-SDR $-6.1$\,dB. Against $R_3$ the reward rises as much ($+0.269$) with PESQ
intact ($-0.034$), and $R_2$ behaves the same ($+0.378$, $-0.034$), so this is a property of patching rather than
of one round; the random-pair baseline again matches it ($+0.019$). Preregistered H3 is \emph{not} supported (2.33 degraded measures for
both rewards): NISQA and UTMOS \emph{rise} and DNSMOS falls under every reward, so none detects this hack, while
speaker similarity and WER barely move. $R_3$ scores the $R_0$-hacked enhancer $2.77$ against $3.72$ for the pre-trained model, flagging a
generator that hacked a \emph{different} predictor but not the one that hacked itself ($3.99$); without a
loudness-matched control we cannot separate this from ``scores degraded audio lower''.

% \section{Limitations}
% \label{sec:limitations}
% Our attacker is black-box and mild: against SSL-MOS its flagged hacks move ESTOI by only $-0.023$ and SI-SDR by
% $-1.034$\,dB, and two independent judges call them perceptually marginal (UTMOS $-0.004$, DNSMOS OVRL $-0.048$),
% whereas PGD examples move UTMOS by $+0.802$. The detector is not fully independent: two panel measures are neural and attackable in principle,
% and for waveform attacks the panel reference is the input itself, so ASR is not comparable across attack
% families. The patch study uses $n=3$ predictor seeds and one predictor family.

\section{Conclusion}
How hackable a speech quality predictor is depends on two measurement choices the field has not standardised:
whether the hack rule is referenced to the processing chain's own output, and whether hackability is reported for
one attacker or the worst of several. Corrected on both, common predictors differ by more than an order of
magnitude at one perturbation size --- NISQA $0.90$ against UTMOS $0.06$ --- so the choice of reward model is
itself a robustness decision. Patching a predictor on its own hacks hardens only the space it trains on, and a
random-perturbation baseline matches it.

Two caveats bound these conclusions. Our attacker is black-box and mild --- its hacks against SSL-MOS move
ESTOI by $-0.023$ and SI-SDR by $-1.034$\,dB, and two independent judges call them perceptually marginal (UTMOS
$-0.004$, DNSMOS $-0.048$; PGD moves UTMOS $+0.802$) --- and we ran no listening test, so the
benchmark ranks predictors against \emph{this} attack family only. The detector is not fully independent either:
two panel measures are neural, and for waveform attacks its reference is the input itself, so ASR is not
comparable across families. The patch study uses $n=3$ seeds and one predictor family (46 A100-hours).

\bibliographystyle{IEEEtran}
\bibliography{refs}
\end{document}